\documentclass[11pt,a4paper]{article}
\usepackage[hyperref]{acl2020}
\usepackage{times}
\usepackage{latexsym}

\usepackage[utf8]{inputenc} % allow utf-8 input
\usepackage[T1]{fontenc}    % use 8-bit T1 fonts
\usepackage{hyperref}       % hyperlinks
\usepackage{url}            % simple URL typesetting
\usepackage{booktabs}       % professional-quality tables
\usepackage{amsfonts}       % blackboard math symbols
\usepackage{nicefrac}       % compact symbols for 1/2, etc.
\usepackage{graphicx}
\graphicspath{{figures/}}
\usepackage{multirow}
\usepackage{hyperref}
\usepackage{enumitem}

\usepackage{microtype}

\aclfinalcopy % Uncomment this line for the final submission
\usepackage{array}
\newcolumntype{P}[1]{>{\centering\arraybackslash}p{#1}}

\title{Revisiting the Open-Domain Question Answering Pipeline}

\author{Sina J. Semnani \\
  Stanford University \\
  \texttt{sinaj@stanford.edu} \\\And
  Manish Pandey \\
  Carnegie Mellon University \\
  \texttt{mpandey@andrew.cmu.edu} \\}

\date{}

\begin{document}

\maketitle

\begin{abstract}

  Open-domain question answering (QA) is the tasl of identifying answers to natural
  questions from a large corpus of documents. The typical open-domain QA system
  starts with
  information retrieval to select a subset of documents from the
  corpus, which are then processed by a machine reader to select
  the answer spans. This paper describes Mindstone, an open-domain
  QA system that consists of a new multi-stage pipeline that employs
  a traditional BM25-based information retriever, RM3-based neural
  relevance feedback, neural ranker, and a machine reading
  comprehension stage. This paper establishes a new baseline for end-to-end performance on question
 answering for Wikipedia/SQuAD dataset (EM=58.1, F1=65.8), with
 substantial gains over the previous state of the art (\citealp{bertserini-plus}). We also
 show how the new pipeline enables the use of low-resolution labels, and can be easily tuned to meet various timing requirements.

% The set of techniques
% presented establish a new baseline for end-to-end performance on question
%answering for Wikipedia/SQuAD dataset (EM=58.1, F1=65.8), an 8 points
%gain over previous baseline (\citealp{bertserini-plus}), while
% achieving a lower end-to-end latency.

\end{abstract}

\section{Introduction\footnote{Note: This work was completed in
summer of 2019. Recent work, including \cite{learning_ret, dpr} have published results
with SQuAD Open EM/F1 scores closer to ours.}}

In this paper, we introduce Mindstone, an open-domain question answering (QA) system,
which answers user's questions from a large collection of documents. We
present our results for QA from Wikipedia using questions from SQuAD~\cite{rajpurkar2016squad}. 
One of the first significant open-domain QA systems was DrQA~\cite{drqa} that combined term-based information retrieval
techniques with a multi-layer RNN-based reader to identify answers in Wikipedia articles. In a more recent work, BERTserini~\cite{bertserini}
replaced the retriever with Anserini~\cite{anserini} and the reader with BERT~\cite{bert}, to obtain large improvements over prior results.
%A more recent followup~\cite{bertserini-plust} uses distant supervision to perform data augmentation.
While BERT and other pre-trained transformer models have enabled machine comprehension
systems to reach human-level performance on many paragraph-level datasets (e.g. the first place on the SQuAD~\cite{rajpurkar2016squad} leaderboard achieves EM/F1=90.0/92.4~\citealp{squadleaderboard}), the overall
performance of open-domain QA with retriever-reader pipelines remains about half of these numbers.
%This prompted us to look at the entire QA pipeline and investigate ways to
%improve the end-to-end performance of the QA systems.

We propose an improved pipeline structure that has enabled Mindstone achieve a
new state-of-the-art QA pipeline peformance on Wikipedia by a significant margin. Our
experiments show that adding two specialized stages for ranking retrieval results and question expansion
improves the end-to-end performance by 8 points, improves answer recall, and makes it easier to use
larger datasets with lower-resolution labels (potentially gathered from users' interaction with the QA system).
We also measure the end-to-end time per query for our system and compare it to prior work.

%In the rest of the paper, we discuss background and related work in
%Section~\ref{gen_inst}. Section~\ref{sec:data} describes the
%specialized data warehousing related QA dataset we use for our
%experiments. Section~\ref{approach} describes the key components of
%the Mindstone QA pipeline. The experimental setup and results are
%desribed in Sections~\ref{experiments} and \ref{results},
%respectively.

%\input{background}
\vspace{-0.1in}
\section{Background and Related Work}
\label{gen_inst}
\vspace{-0.1in}

The machine reading task of answering questions
has made great progress in recent
years. There are two primary reasons for this. The first is the creation
of datasets such as  QACNN/DailyMail~\cite{hermann2015},
%WikiQA~\cite{yang2015wikiqa},
SQuAD~\cite{rajpurkar2016squad}, and Natural
Questions~\cite{nq}.
%and the QAngaroo
%comprehension database~\cite{welbl2018}.
The second is the considerable progress in deep learning architectures like
attention-based and memory augmented neural
networks~\cite{bahdanau2016, weston2014memory} ,
Transformers~\cite{vaswani2017attention} and pre-trained 
models~\cite{bert}. 
 
Until recently, open-domain QA has been mostly
addressed through the task of answering from structured knowledge bases
such as
%WebQuestions~\cite{berant2013} and
Simple
Questions~\cite{bordes2015}. However, KB limitations such as incomplete or
missing information and fixed schemas, have generated new interest in question
answering from unstructured documents.
%Datasets such as MS~MARCO~\cite{msmarco} contain questions sampled
%from real anonymized user queries paired with real web documents
%from the Bing search engine. 

DrQA, the pioneering work of ~\citet{drqa},
answers questions from the entire Wikipedia. Its pipeline
combines a document retriever and a bidirectional
RNN document reader. 
BERTserini \cite{bertserini} uses a paragraph-level Anserini-based
retriever and a fine-tuned BERT reader for answering questions.
In a follow-on work, ~\citet{bertserini-plus} discuss data augmentation
 using distant supervision. 
While this work established the previous best
results on open-domain QA, our experiments show that adding a neural ranker
and neural RM3 to the pipeline results in a faster and more accurate system.

%BERT (Devlin et al., 2018), the latest refinement
%of a series of neural models that make heavy use
%of pretraining (Peters et al., 2018; Radford et al.,
%2018), has led to impressive gains in many natural
%language processing tasks, ranging from sentence
%classification to question answering to sequence
%labeling. Most relevant to our task, Nogueira
%and Cho (2019) showed impressive gains in using BERT for query-based passage reranking. In
%this demonstration, we integrate BERT with the
%open-source Anserini IR toolkit to create BERTserini, an end-to-end open-domain question answering (QA) system.
%Unlike most QA or reading comprehension
%models, which are best described as rerankers or
%extractors since they assume as input relatively
%small amounts of text (an article, top k sentences
%or passages, etc.), our system operates directly on
%a large corpus of Wikipedia articles. We integrate
%best practices from the information retrieval community with BERT to produce an end-to-end system, and experiments on a standard benchmark
%equal contribution
%test collection show large improvements over previous work. Our results show that fine-tuning
%pretrained BERT with SQuAD (Rajpurkar et al.,
%2016) is sufficient to achieve high accuracy in
%identifying answer spans. The simplicity of this
%design is one major feature of our architecture. We
%have deployed BERTserini as a chatbot that users
%can interact with on diverse platforms, from laptops to mobile phones.

%\input{data}

%\input{pipeline}
\section{Pipeline Architecture}
\label{approach}

\begin{figure*}
  \centering
  \includegraphics[width=0.5\textwidth, trim={10 110 20 110}, clip]{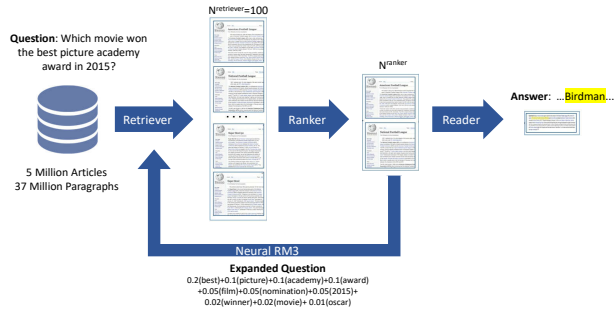}
  \caption{Mindstone pipeline architecture. Numbers are for the Wikipedia corpus.}
  \label{fig:pipeline}
\end{figure*}

% \subsection{Mindstone pipeline architecture}
In this section, we describe the Mindstone pipeline.
Following~\citet{bertserini}, we first split the corpus articles into paragraphs.
We also prepend article titles to each paragraph to provide some context from the full article.
We will use paragraphs and documents interchangeably to refer to the resulting paragraphs.
For each question,
all documents travel through all stages of the pipeline or until it is apparent that they cannot score high enough to be included in top answers.
Figure~\ref{fig:pipeline} depicts these stages and their relative order.

Assume the corpus has $N^{corpus}$ paragraphs.
Given a question, retriever, ranker and reader each
assign a score to every text span of every paragraph.
We use $S^{retriever}$, $S^{ranker}$
and $S^{reader}$ to denote these scores.
In practice, all text spans in the same paragraph have the same retriever and ranker scores,
and we only calculate $S^{ranker}$ for top $N^{retriever}$ documents (when sorted by retriever score)
to lighten the job of the slower ranker.
The final score of a span of text is a weighted average of
its three scores, where the weights are tuned to maximize the exact match metric
on a small subset of the training set.
We normalize all scores to have a value in $(-\infty, 1]$ before taking the average.
%MP - redacted for public version of paper  
% Having an upper bound allows us to detect documents that cannot have a top answer span due to their low
%$S^{retriever}$ or $S^{ranker}$, and
%prevent the reader from processing them.
%As a result, in our experiments on average only about $2.5\%$ of documents go through the reader.
%This way, we can use very large neural networks as reader without slowing down the pipeline.

\paragraph{Retriever.}
We use a TF-IDF-based retriever to retrieve $N^{retriever}$ documents.
More specifically, we use Anserini~\cite{anserini} based on Lucene version $8.0$,
and Okapi BM25.
Our index only considers unigrams that are not in a predefined set of stop words.\footnote{Adding bigrams for a paragraph-level corpus slightly hurts the pipeline's performance and speed.
DrQA uses unigrams \emph{and} bigrams for indexing, which improves the performance on an article-level corpus.}
The main advantage of this stage is its speed, and it needs to have a high recall$@N^{retriever}$ since
the performance of the full pipeline is upper bounded by it.

\paragraph{Ranker.}
We use a BERT-Base ($110M$ parameters) model for ranking the retrieved documents.
The model is trained as a binary classifier on MS~MARCO~\cite{msmarco} using the same method as~\citet{bert-reranking}, and then fine-tuned
on SQuAD. For more details on our fine-tuning approaches, see section \ref{experiments}.
$S^{ranker}$ is the output of the classifier before softmax layer. Our ranker takes the first 448 tokens of the paragraph
and the whole question as input and outputs whether the paragraph contains an answer to the question.

%MP - redacted for public version of paper  
%We believe adding a ranker and using MS~MARCO to train it helps with generalizing to other datasets (see section \ref{results}).

\paragraph{Neural RM3.}
Our experiments showed that using RM3~\cite{rm3} in retriever hurts its recall.
Instead, we use a method similar to RM3, but using ranker document scores instead of retriever scores.
Let $d_1, ..., d_{N^{retriever}}$ denote the sorted output of retriever and
$S^{ranker}_1, \geq ... \geq S^{ranker}_{N^{retriever}}$ be the scores that ranker
has assigned to them. Let $v(d)$ be the vector of TF-IDF scores of the top $T$ most common terms in
document $d$ where $T$ is a hyperparameter. Then
\[
  q'=\alpha q + (1-\alpha) \Sigma_{S^{ranker}_i>0} v(d_i)
  \]
where $q$ is the original question vector and $\alpha \in [0, 1]$ is a hyperparameter, will be the expanded question in vector form.
We use this new question to retrieve more documents from the corpus and feed them through the ranker.
In the open-domain Wikipedia/SQuAD setting, neural RM3 increases recall$@100$ by $6$ points.
% improves the exact match of the full pipeline only by $0.5$ points at the expense of a slowdown.
However, due to the slowdown it causes, in section \ref{results} we report our results without neural RM3.
% We believe this method can be especially useful for multi-hop question answering, and leave it to future work.

\paragraph{Reader.}
The reader finds the exact location of the answer in the ranked documents. We assume the answer is a contiguous span of text and use a linear layer for two token-level classification tasks to determine the start and the end position of the answer span~\cite{bert}.
Paragraph and question are truncated or padded to be exactly 384 tokens in total.
Since we are using version 1.1 of SQuAD, the reader always returns an
answer. We experiment with both base and large BERT models.

\section{Experiments}
\label{experiments}

\paragraph{Data.}

Similar to DrQA~\cite{drqa}, and BERTserini~\cite{bertserini-plus}, we use 2016-12-21 English Wikipedia dump, which has more than
5 million articles and 37 million paragraphs.

We use SQuAD version 1.1 to train the
paragraph reader and MS~MARCO to train the ranker.
We have trained all models on the training set of the mentioned datasets.
Since SQuAD's test set is not publicly available, we use a small subset of its training set for development,
and report the results on its development set.

\paragraph{Training Ranker.}
\label{fine-tuning-ranker}
We train the ranker on MS~MARCO, and then fine-tune it on a SQuAD-based
binary classification dataset.
We experimented with three different approaches for building this dataset:

\begin{enumerate}[noitemsep,topsep=0pt]
    \item (Fine-tuning)
    For every paragraph-question pair in SQuAD, 
    add another paragraph from the same Wikipedia article that does not contain the answer string.
    
    \item (Data augmentation 1)
    Use the retriever to retrieve $n$ documents for each question in the dataset, and add them to
    the new dataset. Their labels are determined according to whether or not they include the answer string.
    
    \item (Data augmentation 2)
    Similar to the second approach, use the retriever to obtain $m$ paragraphs, then
    rank them with the ranker and use the top $n$ results.
\end{enumerate}
We use $m=100$ and $n=5$ for our experiments.

% \paragraph{Training Reader.}
% We use the same approach as~\citet{bert}. We train for 2 epochs with
% learning rate $3\cdot 10^{-5}$ and batch size of $18$.

% For Snowflake, we use query generation to make up for the small size of the dataset.
% We start from the reader trained on SQuAD, then fine-tune it on the generated dataset first.
% A final fine-tuning is done on the Snowflake training set. This multi-step approach is more effective than
% training on a mixture of generated and original queries.
% To prevent over-fitting, we freeze every third layer of BERT during training, and change
% which layers are frozen after each epoch. We use learning rate $10^{-6}$ and train for 50 epochs
% with batch size $18$.

%\input{results}
\section{Results}
\label{results}

Unless stated otherwise, we use the same metrics (exact match, F1 and recall) as defined in~\citet{drqa}.
$N^{retriever}$ is set to 100 to be comparable to prior
work.

\subsection{Full Pipeline Performance}

In this section, times are measured on a machine with a single NVIDIA V100 GPU (16GB memory) and an 8-core CPU.
Mixed precision\footnote{\url{https://github.com/NVIDIA/apex}} is used for inference.
All time measurements are averaged over 1000 queries in batch mode, and the minimum of 5 runs is reported.
To have a fair comparison, we have partially re-implemented DrQA to use Anserini, which improves its speed.
We have implemented~\citet{bertserini} and~\citet{bertserini-plus} following their descriptions in the cited papers,
and use our implementations to measure time. Exact match and F1, however,
are directly copied from~\citet{drqa},~\citet{bertserini} and~\citet{bertserini-plus}.

% \subsubsection{Open-domain QA with Wikipedia and SQuAD}
Our main results for open-domain QA from Wikipedia using SQuAD questions are shown
in Table \ref{wiki-pipeline}.
Our best-performing system has a BERT-Large reader.
The ranker training approach that resulted in the best performance is 
fine-tuning followed by data augmentation with neural ranker (the first and third approach from section~\ref{fine-tuning-ranker}).

Although DrQA uses a much smaller neural network, its end-to-end time per query is greater than
Mindstone due to its larger index (DrQA uses unigrams \emph{and} bigrams) and reader's named entity recognition.
BERTSerini and~\citet{bertserini-plus} are slower than Mindstone due to
the additional time spent on
reader's token-level classification and processing multiple segments
for documents that are longer than BERT's maximum sequence length.
Mindstone on the other hand, only reads $2.5\%$ of documents and its ranker only processes the truncated version of long paragraphs. 

One of our findings is that adding a ranker enables a better use of larger reader models.
In previous approaches such as~\citet{bertserini},
using a BERT-Large reader would increase query time by $63\%$, while in Mindstone it increases by only $2\%$.
In addition, ranking is an easier task for a neural network to learn and requires less supervision.
In particular, we were able to use the low resolution labels of a larger and more diverse dataset (MS~MARCO) to train our ranker
because ranking does not require answers to be known exactly.
Compared to reader, it is more straightforward to build datasets for neural ranking
from users' clicks and other interactions.

\begin{table}
  \centering
  \begin{small}
    \begin{tabular}{lccP{0.12\columnwidth}}
      \toprule
       Model    & EM  & F1 & Time per query \\
      \midrule
      DrQA~\cite{drqa}     & 29.8 & -     &  988 ms\\
      BERTSerini~\cite{bertserini}     & 38.6       & 46.1 & 887 ms\\
      \citet{bertserini-plus}     & 50.2       & 58.2 & 887 ms\\
      Mindstone (ours)     & \bf{58.1}       & \bf{65.8} & 738 ms\\
      \bottomrule
      \vspace{-15pt}
    \end{tabular}
    \caption{Open-domain results for SQuAD development set and Wikipedia. DrQA retrieves 5 articles, and all other systems retrieve 100 paragraphs.}
    \label{wiki-pipeline}
  \end{small}
\end{table}

Table~\ref{wiki-ablation} shows the results of our ablation study.
Note that even Mindstone with a BERT-Base reader outperforms the previous state-of-the-art.

\begin{table}
  \centering
  \begin{small}
    \begin{tabular}{lccP{0.15\columnwidth}}
      \toprule
      Model    & EM & F1  & Time per query  \\
      \midrule
      Mindstone (ours)     & \bf{58.1}       & \bf{65.8} & 738 ms\\
      ~- BERT-Large reader     & 53.9 & 62.7 & 722 ms\\
      ~- ranker data augmentation     & 47.8 & 56.4 & \\
      ~- ranker fine-tuning & 44.3 & 53.7 & \\
      \bottomrule
    \end{tabular}
    \caption{The effect of different parts of Mindstone.}
    \label{wiki-ablation}
  \end{small}
\end{table}

\begin{figure}
  \centering
  \includegraphics[width=0.9\columnwidth, trim={100 80 70 50}, clip]{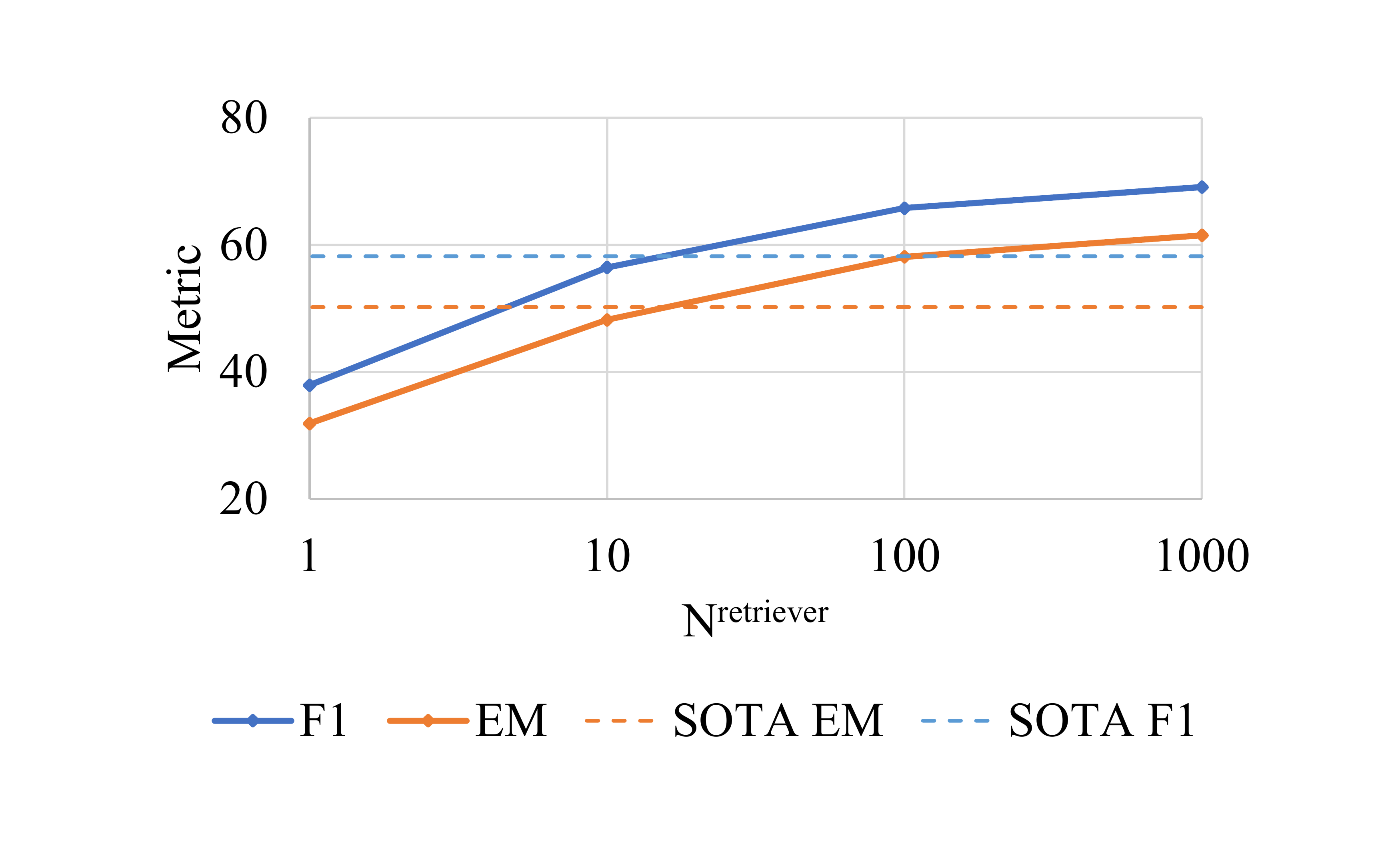}
  \caption{Performance improves with more retrieved documents.}
  \label{fig:tradeoff}
\end{figure}

Figure \ref{fig:tradeoff} shows that unlike DrQA (as shown in~\citealp{weaver}), 
there is a trade-off between speed and accuracy that can be leveraged according to
the timing requirements of the application.
Mindstone surpasses the previous state-of-the-art pipeline by processing 5x fewer paragraphs (at $N^{retriever}=20$), which is about 5x faster.

\subsection{Retriever and Ranker Performance}
\begin{figure}
\centering
\includegraphics[width=0.9\columnwidth, trim={12 100 8 90}, clip]{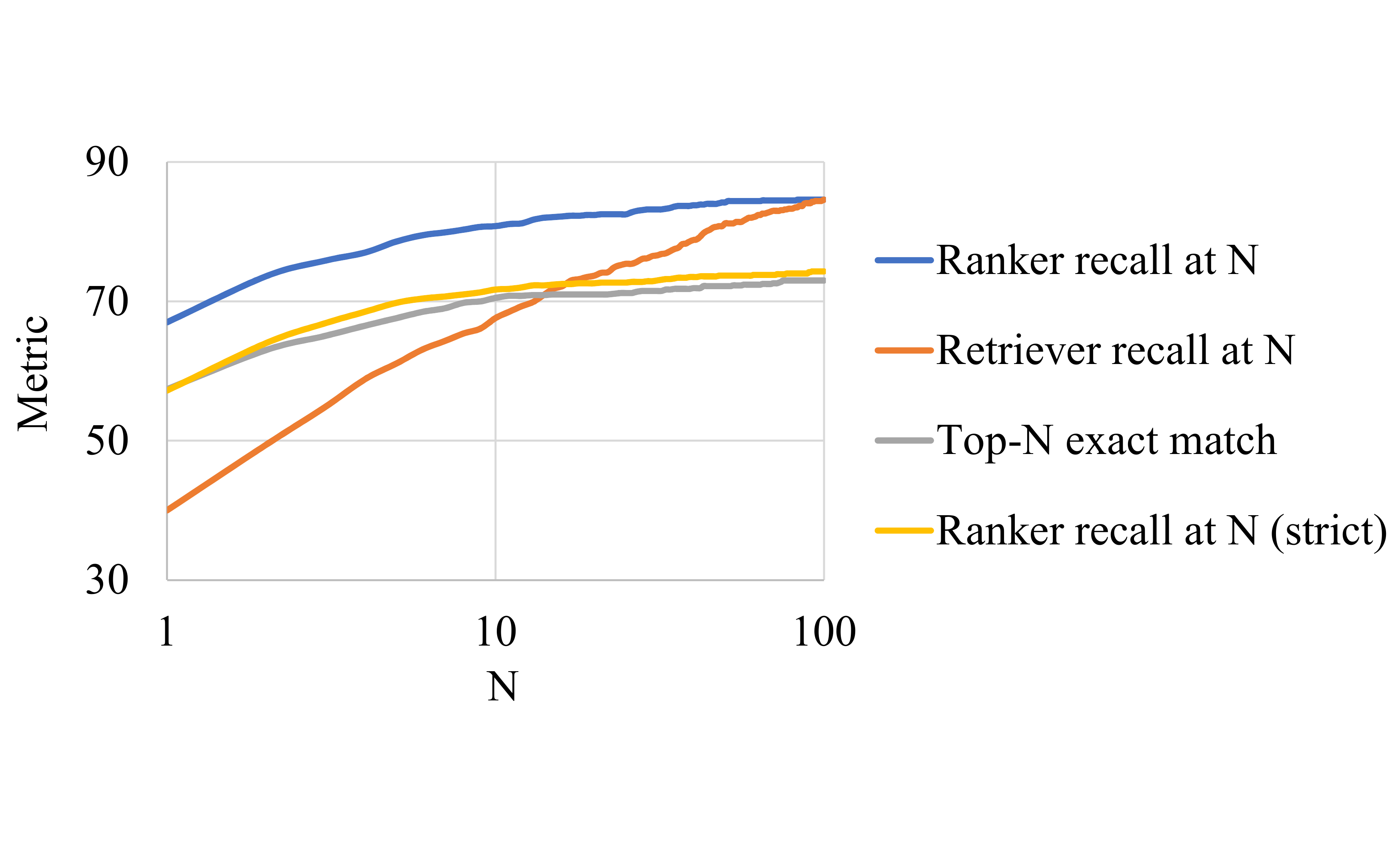}
\caption{Retriever and ranker recall. The x-axis is log-scale.}
\label{fig:recall}
\end{figure}

In Figure \ref{fig:recall}, retriever and ranker recalls show how much ranker improves the retrieval process.

Mindstone outputs a sorted list of answers, so we can calculate accumulative exact match and F1.
Top-N exact match in Figure \ref{fig:recall} shows the probability of finding at least one exact-match answer
in the top $N$ answers.

We also analyze the performance gap between retriever and the full pipeline.
Prior work and other sections of this paper all use the exact match metric in the same way:
if a paragraph includes the gold answer string, it is a hit in terms of recall.
However, this approach can overestimate recall for queries that have
a common phrase as their answer. As a more strict approach, we consider
a retrieved paragraph to be a hit only if it is the same as the paragraph that
crowdworkers used to write the query in the process of building SQuAD.~\footnote{Since the version of Wikipedia dump we are using is different from that of SQuAD's (and other processing differences),
a simple string equality check would not work. Our solution is using a simple string similarity metric and a threshold.}
Using this definition, Figure \ref{fig:recall} shows that the top-N exact match
and recall at $N$ curves are almost identical. This means that Mindstone's reader is not the 
performance bottleneck, and future attempts to improve the pipeline need to be focused on the retriever or ranker.

\section{Conclusion and Future Work}

Our system uses a conventional retriever supplemented with a
neural ranker and neural RM3 relevance feedback.
Our new pipeline design establishes new state-of-the-art results
for end-to-end performance on question
answering for Wikipedia/SQuAD dataset, with an 8 points
gain over previous baseline~\cite{bertserini-plus}.
% Second, the adaption
% techniques demonstrate large performance gains (EM and F1 gains of more than 10
% points), over a system that has not been domain adapted.

% Additionally, we have demonstrated large gains in question answering
% performance that can be achieved using domain adaptation
% techniques described in the paper. Further improving the question generation
% component is one direction for future work.

We have built a highly responsive QA system with a sub-second
latency. While conventional retrievers can
operate with a small latency, the computationally heavy ranking and
reader stages that use BERT, can slow down the pipeline.
This can be mitigated by tuning the number of
documents retrieved and processed by ranker and reader.
Techniques such as Distillation~\cite{distilling} that reduce the
model size
%, and newer models such as ALBERT~\citet{anonymous2020albert} that have
%fewer parameters
can play a role in increasing the speed of QA systems.

\medskip
\small

\bibliographystyle{acl_natbib}
\bibliography{bibliography}

\end{document}